\documentclass[letterpaper, 10pt, conference]{ieeeconf}  
\usepackage{cite}
\usepackage[utf8]{inputenc}
\IEEEoverridecommandlockouts                             
\usepackage{amsmath} 
\let\labelindent\relax
\usepackage{enumitem}
\usepackage{graphicx}
\usepackage{subfig} 
\usepackage{booktabs}
\usepackage[hidelinks]{hyperref}            
\usepackage{float}
\usepackage{siunitx}
\usepackage{overpic}
\usepackage{multirow}
\usepackage[table]{xcolor}
\usepackage{caption}
\usepackage{placeins}    
\usepackage{calc}        
\usepackage[absolute]{textpos}  
\usepackage{stfloats}  
\usepackage{cuted} 
\usepackage{caption}
\usepackage{censor}
\usepackage{amssymb}

\title{\vspace{-12pt}\bf PhysiFlow: Physics-Aware Humanoid Whole-Body VLA via Multi-Brain Latent Flow Matching and Robust Tracking}
\author{Weikai~Qin$^{*}$, Sichen~Wu$^{*}$, Ci~Chen, Mengfan~Liu, Linxi~Feng, Xinru~Cui, Haoqi~Han, Hesheng~Wang$^{\dag}$
\thanks{Weikai~Qin, Sichen~Wu, Ci~Chen, Mengfan~Liu, Linxi~Feng, Xinru~Cui and Haoqi~Han is with the department of Automation Engineering, 
        Shanghai Jiao Tong University, Shanghai, 200240, China 
        (e-mail: LingMisaki@sjtu.edu.cn, 15182340716@163.com, 21813007@zju.edu.cn, mengfan\_liu@sjtu.edu.cn, linxif2008@sjtu.edu.cn, cxr0726@sjtu.edu.cn and hhq123@sjtu.edu.cn)}
\thanks{$^{*}$Equal contribution} 
\thanks{$^{\dag}$Hesheng Wang is with the Department of Automation, Key Laboratory of System Control and Information Processing of Ministry of Education, Key Laboratory of Marine Intelligent Equipment and System of Ministry of Education, Shanghai Engineering Research Center of Intelligent Control and Management, Shanghai Jiao Tong University, Shanghai, 200240, China (e-mail: wanghesheng@sjtu.edu.cn)}
}
\usepackage{capt-of}
\usepackage{graphicx}

\makeatletter
\let\@oldmaketitle\@maketitle
\newif\if@titlefigure@done
\@titlefigure@donefalse

\renewcommand{\@maketitle}{
    \@oldmaketitle
    \if@titlefigure@done\else
        \vspace{-6pt} 
        \begin{center}
            \includegraphics[width=1.0\textwidth]{"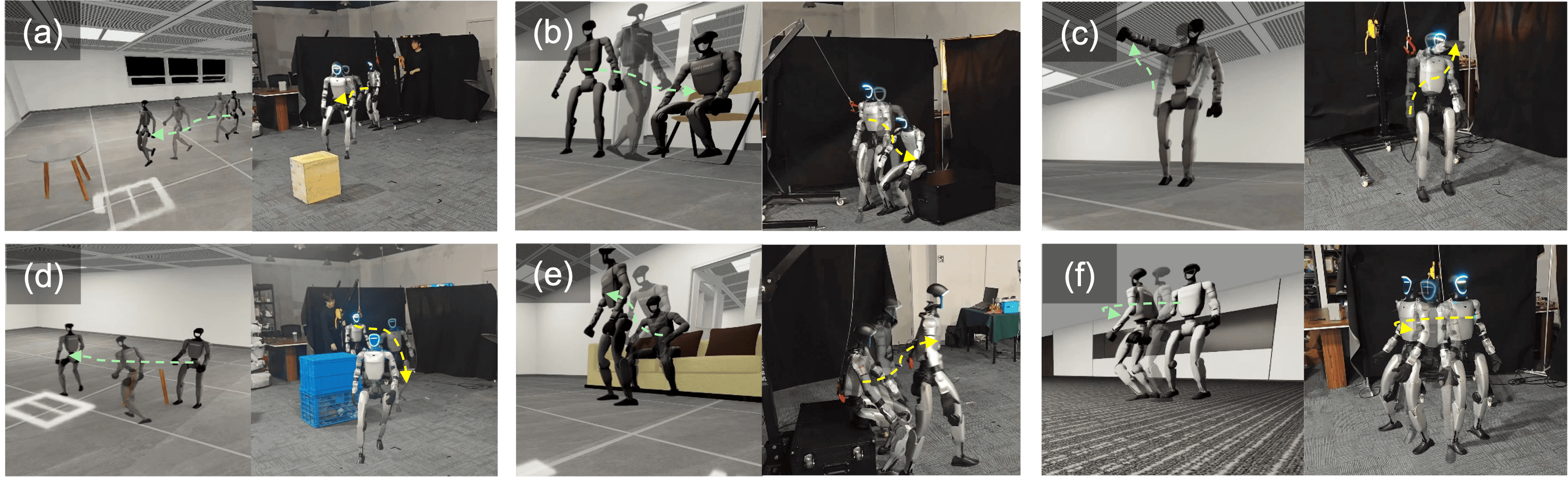"}
            \captionof{figure}{Introducing \textbf{PhysiFlow}, a multi-brain VLA humanoid system that operates on Unitree G1 robots and performs end-to-end VLA humanoid whole body control in large spaces. The proposed system achieves consecutive tasks autonomously, including \textbf{(a-c)} walking to the designated item, sitting on the proposed item, and raising arm; \textbf{(d-f)} circling the designated item, standing up from the specific item and turning right.}
            \label{first picture}
        \end{center}
        \vspace{6pt}
        \@titlefigure@donetrue
    \fi
}
\makeatother

\begin{document}
\maketitle
\begin{abstract}

In the domain of humanoid robot control, the fusion of Vision-Language-Action (VLA) with whole-body control is essential for semantically guided execution of real-world tasks.  However, existing methods encounter challenges in terms of low VLA inference efficiency or an absence of effective semantic guidance for whole-body control, resulting in instability in dynamic limb-coordinated tasks. To bridge this gap, we present a semantic-motion intent guided, physics-aware multi-brain VLA framework for humanoid whole-body control. A series of experiments was conducted to evaluate the performance of the proposed framework. The experimental results demonstrated that the framework enabled reliable vision-language-guided full-body coordination for humanoid robots.

\end{abstract}

\section{INTRODUCTION}
Humanoid robots, which possess a similar physical structure to humans, have the potential to execute a variety of tasks and motions that are prevalent in human daily life.  Nowadays, great efforts have been made to enhance their locomotion and manipulation capabilities: advances in learning-based\cite{cheng2024expressive,fu2024humanplus} and model-based \cite{ramos2019dynamic} whole-body control methodologies have enabled humanoid robots to perform a diverse range of tasks with high flexibility and precision in execution\cite{gu2025humanoid,he2025hover}.  Meanwhile, VLA models have gained widespread application in robotic manipulation\cite{kim2024openvla}. By unifying visual perception, language understanding, and action generation within an end-to-end framework, they eliminate the interface barriers of traditional modular systems\cite{qu2025spatialvla}. Besides, they endow robots with intuitive human-robot interaction and autonomous semantic task understanding capabilities.

The fundamental challenge encountered in the field of humanoid VLA is the integration of semantic perception, high-frequency motion generation, and physics-aware control in existing frameworks. Traditional VLA-driven systems suffer from low inference efficiency and poor adaptability with edge deployment\cite{bjorck2025gr00t,zhang2025mole}, while learning-based whole-body control methods often lack effective guidance from vision-language semantics\cite{ji2024exbody2,wu2025stride}. Consequently, humanoid robots often exhibit instability and task failure in vision-and-language-guided dynamic whole-body control scenarios that demand simultaneous upper and lower limb coordination and balance maintenance\cite{sapkota2025vision}. To address this problem, we present a semantic-motion intent guided, physics-aware multi-brain VLA framework for humanoid whole-body control, which enables reliable full-body motion performance with real-time inference and physical stability via semantic-motion integration.

The key innovation of this study lies in the bio-inspired hierarchical multi-brain architecture, which is centred on the semantic latent vector. This decouples and coordinates three functional brains for the purpose of humanoid whole-body control. The Neocortical Brain fuses task semantics of what to do and motion intent of how to do via a curriculum-based conditional variational autoencoder (CVAE) based on SigLIP\cite{zhai2023sigmoid} with LoRA lightweight adaptation. The Basal Ganglionic Brain generates \qty{50}{\Hz} high-frequency motion sequences through latent vector-driven flow matching. The Cerebellar Brain is a motion tracker that enforces physics-aware constraints via joint fine-tuning and tracking error backpropagation. This design successfully overcomes the conventional conflict between inference efficiency, semantic generalization, and dynamic balance in VLA systems, effectively adapting to full-body tasks guided by vision and language that require upper and lower limb coordination.

The proposed framework is validated on the Unitree G1 humanoid robot through a combination of simulation and real-world experimentation. A series of tests are conducted in order to evaluate whole-body control scenarios. These include the ability to sit on or circle a specific item, which is shown in Fig. \ref{first picture}. In the context of simulation, the proposed method has been shown to outperform existing baseline approaches with regard to task success rate. The reliability of the system is further demonstrated by real-world experiments. The key contributions are summarized as follows:

\begin{enumerate}
    \item A novel bio-inspired multi-brain VLA framework is proposed, decoupling high-level semantic-motion intent inference from low-level high-frequency motion generation and stable tracking control for humanoid whole-body control.
    \item In Neocortical Brain, a two-phase CVAE curriculum based on SigLIP with LoRA lightweight adaptation is employed to generate the modality-invariant semantic latent vector with motion intent.
    \item Driven by latent vector, we introduce a physics-aware flow-matching training paradigm by fusing motion tracking with joint fine-tuning for dynamic and consistent motion generation.
\end{enumerate}

\section{RELATED WORK}

\subsection{VLA for Robotics Learning}
VLA models have become the mainstream of end-to-end robotic learning, which unify multi-modal perception and action generation to break the interface barriers of traditional modular systems \cite{intelligence2025pi_}. For fixed-base dual-arm manipulation, representative works like $\pi_0$ \cite{black2024pi_0} and CO-RFT \cite{kim2025fine} have achieved excellent generalization and trajectory quality, but they pose significant challenges when migrated to humanoid robots: insufficient inference speed for high-frequency full-body control \cite{zhang2025up}, and excessive computing overhead that exceeds on-board edge device limits \cite{zhen20243d}. To solve these issues, recent humanoid VLA works widely adopt dual-brain decoupled architectures, such as Galaxea G0 \cite{jiang2025galaxea} and LeVERB \cite{xue2025leverb}, which effectively improve inference efficiency \cite{song2025rationalvla}. However, existing evaluations are mostly limited to simple tabletop tasks, lacking physics-aware stable control for complex whole-body humanoid motion with coordinated limb movement \cite{cui2025openhelix}. In this work, we propose a multi-brain VLA framework to fill this gap, enabling robust whole-body control for humanoid robots in complex scenarios.

\subsection{Humanoid Whole-Body Control
}
Whole-body control of humanoid robots remains a formidable challenge, due to their highly nonlinear dynamics and multi-joint coordination requirements\cite{cheng2024expressive,fu2024humanplus}. Although traditional model-based approaches\cite{ramos2019dynamic,gu2025humanoid} are effective for basic motions in structured environments, it is often difficult to balance flexibility and stability in complex scenarios. Recent advances in learning-based control, such as RL and teacher-student frameworks, have demonstrated significant progress in flexible motions and multiple tasks\cite{he2025hover,ji2024exbody2,wu2025stride}. TWIST leverages an RL and behaviour cloning (BC) teacher-student pipeline with real-time motion capture (MoCap) retargeting, enabling it to execute coordinated skills like whole-body manipulation and locomotion\cite{ze2025twist}, while ASAP proposes a two-stage framework with a delta action model and asymmetric actor-critic training to compensate sim2real physics mismatch\cite{he2025asap}. However, these methods fundamentally rely on motion tracking and teleoperation, lacking autonomous comprehension of vision input and language guidance\cite{deng2025survey}. The existence of these high-level perceptual decision-making abilities is of critical importance for humanoid robots to transition into domestic service scenarios, which represents the primary gap in current research.
\begin{figure*}[!t]\centering
\vspace{-6pt}
    \includegraphics[width=\linewidth]{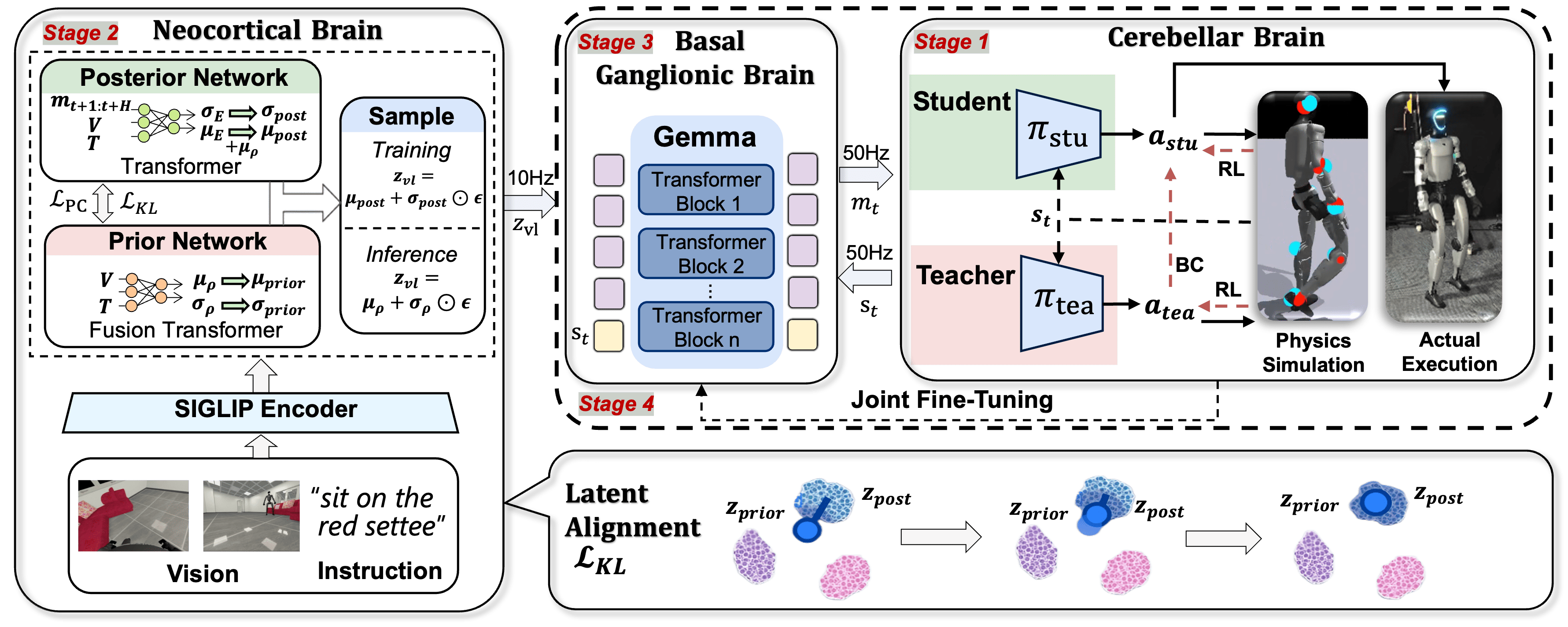}
\caption{\textbf{The overall pipeline of PhysiFlow.} This bio-inspired architecture decouples semantic reasoning from physics-aware execution. \textbf{(a) Neocortical Brain:} A curriculum-based CVAE processes vision and language to synthesize a \qty{10}{\Hz} latent vector $z_{vl}$, aligning task semantics with motion intent. \textbf{(b) Basal Ganglionic Brain:} Conditioned on $z_{vl}$ and robot states, a flow-matching model generates \qty{50}{\Hz} motion sequence $m_t$ for continuity. \textbf{(c) Cerebellar Brain:} A robust motion tracker enforces physical constraints, translating these chunks into stable motor commands for closed-loop whole-body control.}
\vspace{-6pt}
\label{overall_framework}
\end{figure*}

\section{Method}  

\subsection{Multi-brain Architecture}
We propose a bio-inspired hierarchical multi-brain VLA framework which is illustrated in Fig. \ref{overall_framework} for humanoid whole-body control, focusing on semantic-motion intent alignment, efficient motion generation, and physics-aware dynamic action execution. The full framework is described as follows.

\textbf{1) Neocortical Brain: Semantic-Motion Intent Alignment}

The Neocortical Brain Module employs a two-stage curriculum-based CVAE as its VLA prior model. Inspired by LeVERB\cite{xue2025leverb}'s residual CVAE architecture, we introduce the latent vector $z_{vl}$ as the primary training objective. Its core function is to integrate the semantic instruction of what to do with the motion intent of how to do. Furthermore, in the inference stage, it can generate \qty{10}{\Hz} semantic-motion intent $z_{vl}$ directly from current vision input and language guidance alone, without knowing any future motion sequences and the current state, making it ideal for real-time control scenarios. The detailed structural design and training specifics of $z_{vl}$ are elaborated on in subsequent sections.

\textbf{2) Basal Ganglionic Brain: Semantic-Motion Intent Driven Flow Matching}

Inspired by the flow matching mechanism in $\pi_{0.5}$ \cite{intelligence2025pi_}, this module innovatively replaces traditional direct vision-language encoders with the \qty{10}{\Hz} latent vector $z_{vl}$,  which is trained in Neocortical Brain. Along  with the robot's current state, it forms the input to the flow-matching model, enabling continuous motion sequence generation guided by both semantic and motion-trend intent information. This approach ensures generated motion sequences not only meet task semantic requirements, but also inherit logical consistency and continuity in motion naturally. The model uses the low-parameter action expert network Gemma\cite{team2024gemma}, which leverages the capability of flow matching to generate high-frequency motion sequence chunks. This ensures \qty{50}{\Hz} real-time inference performance, which is ideal for the high-frequency feedback demands of full-body control in humanoid robots. During training, the model learns flow field mappings from noisy motion sequences to real motion sequences. It uses a beta distribution time-step sampling strategy ($\alpha$ = 1.5, $\beta$ = 1) to enhance training stability. 

\textbf{3) Cerebellar Brain: Motion Tracking with Joint Fine-Tuning}

The motion tracker is used in this module to achieve stable, high-precision tracking of input motion sequences. The pre-trained motion tracking model is optimised using a combination of RL and teacher-student learning strategies, while domain randomisation enhances sim2real capabilities. During deployment, the motion tracking module receives \qty{50}{\Hz} motion sequence inputs. These are converted by an internally running \qty{1000}{\Hz} PD controller into executable motor commands for the Unitree G1 humanoid robot, achieving closed-loop control from semantic intent to physical action $a_t$.

During training, a sequential optimization pipeline is employed to ensure physically viable, high-frequency control. As illustrated in Fig. \ref{overall_framework}, the process is divided into four distinct stages. In stage 1, we first pre-train the Cerebellar Brain on a physics-informed motion dataset to build a robust motion tracker. Then, in stage 2, the Neocortical Brain CVAE is trained via two-phase curriculum learning with staged loss and LoRA adaptation. For the stage 3, the Basal Ganglionic flow-matching module is trained with a Gemma\cite{team2024gemma} decoder through sampling, path building, and the MSE loss minimization process. Finally in stage 4, we propose a joint fine-tuning strategy for the flow matching model and the frozen motion tracking module. The error between executed and reference actions from the two sequential modules is backpropagated to fine-tune the flow matching model, ensuring its generated motions are physics-aware and dynamically consistent with motion tracking constraints. This hierarchical paradigm dynamically bridges abstract semantic reasoning with stable physical execution.

\subsection{The design of the Neocortical Brain}

To achieve multi-modal semantic alignment across vision, language, and motion intent, we build our model upon a CVAE framework with dedicated feature encoding and fusion modules. The vision encoder employs a pre-trained SigLIP ViT-B/16 model, which processes images from the first-person view $V^{ego}_t$ and the third-person view $V^{exo}_t$ to comprehensively capture scene information. The text encoder also originates from SigLIP, leveraging its contrastive pre-training advantage to encode text instructions $T$ into semantically consistent textual features. The two feature types are further integrated via a Fusion Transformer to generate a unified vision-language representation, which is projected to parameterize the prior and posterior distributions of the latent verb \(z_{vl}\).

Inspired by the residual CVAE design in LeVERB \cite{xue2025leverb}, we construct the prior and posterior networks as follows.
The prior network \(p_\rho(\mu_{\rho},\sigma^2_{\rho}|V,T)\) of the CVAE outputs the mean \(\mu_\rho\) and diagonal variance \(\sigma^2_\rho\), modeling scene semantics and task objectives, where $\mu_{prior}=\mu_{\rho}$, $\sigma^2_{prior}=\sigma^2_{\rho}$. The posterior network \(q_E(\mu_{E},\sigma^2_{E}|V,T,m_{t+1:t+H})\) additionally takes the privileged future motion sequences \(m_{t+1:t+H}\) (only accessible during training) as input, predicting mean \(\mu_E\) and diagonal variance \(\sigma^2_E\) to capture motion intent details that cannot be inferred from vision-language inputs alone.

During training, we construct the posterior distribution via residual fusion to decouple semantic encoding and motion detail completion: the posterior variance $\sigma^2_{post}=\sigma^2_{E}$ is directly inherited from the posterior network, and the posterior mean is defined as \(\mu_{post} = \mu_\rho +\mu_E\). The standard reparameterization trick is applied for gradient backpropagation with stage-specific sampling rules: in the training stage, the 256-dimensional latent verb $z_{vl}$ is sampled from the posterior distribution as \(z_{vl} = \mu_{post} + \sigma_{post} \odot \epsilon\) to leverage privileged motion information, where \(\epsilon \sim \mathcal{N}(0,I)\). At inference, with no access to future motion sequences, we sample $z_{vl}$ directly from the prior network’s output distribution as \(z_{vl} = \mu_{\rho} + \sigma_{\rho} \odot \epsilon\).

The total loss function focuses on core CVAE optimization objectives while integrating our proposed semantic alignment and regularization terms, expressed as 
\begin{equation}
\mathcal{L}=\mathcal{L}_{Recon} + \beta \mathcal{L}_{KL} + \mathcal{L}_{PC} + \mathcal{L}_{VL} + \mathcal{L}_{Aux}+\lambda_{Disc}\mathcal{L}_{Disc}
\end{equation}

The specific definition of each loss term is as follows:
\begin{equation}
    \mathcal{L}_{Recon} = \sum_{t=1}^H w_t \cdot \|\hat{ m}_t -  m_t\|_2^2
\end{equation}
\begin{equation}
\mathcal{L}_{KL}=D_{KL}\big(\mathcal{N}(\mu_{post},\sigma_{post}^2)\ \|\ \mathcal{N}(\mu_\rho,\sigma_\rho^2)\big)
\end{equation}
\begin{equation}
   \mathcal{L}_{PC}=\sum_{t=1}^H\|\hat{m}^{\mu_{prior}}_t-\hat m^{\mu_{post}}_t\|_2^2
\end{equation}
\begin{equation}
\mathcal{L}_{VL}=-\log\frac{\exp(\cos(\mu_\rho,t^+)/\tau)}{\sum_j \exp(\cos(\mu_\rho,t_j)/\tau)}
\end{equation}
where $\mathcal{L}_{Recon}$ is the time‑step weighted reconstruction loss on the motion increment $ m_t$, emphasizing short‑term motion intent. $\mathcal{L}_{\text{KL}}$ aligns the posterior $\mathcal{N}(\mu_{\text{post}}, \sigma_{post}^2)$ with the prior $\mathcal{N}(\mu_\rho, \sigma_\rho^2)$ to regularize latent space, ensuring the prior network approximates the posterior's motion encoding capability for inference. $\mathcal{L}_{PC}$ distills posterior knowledge by aligning $\hat{m}^{\mu_{prior}}_t$ and $\hat{m}^{\mu_{post}}_t$, enabling the prior to mimic the posterior’s motion intent at inference without access to future sequences, where $\hat{m}^{\mu_{prior}}_t$ is motion reconstructed from the prior mean $\mu_\rho$ under inference-like conditions, and $\hat{m}^{\mu_{post}}_t$ is motion reconstructed from the posterior mean $\mu_{post}$ with privileged future information. $\mathcal{L}_{VL}$ applies InfoNCE alignment between $\mu_\rho $ and text embeddings to bind the latent prior to language semantics and improve inference-time retrieval performance, while the $t^+$ is the matched text embedding for the current sample, $t_j$ denotes the candidate text embeddings in the contrastive set.  $\mathcal{L}_{Disc }$ is a modality‑adversarial loss with a Gradient Reversal Layer (GRL) that enforces modality invariance of $z_{vl}$ for enhanced cross-modal robustness, which is a mature paradigm in the field and thus no formulation is needed.  $\mathcal{L}_{Aux}$ aggregates auxiliary objectives including instruction classification, z‑use margins, $\sigma_\rho $ regularization, which are all basic optimization terms and no separate formulation is needed. 

We adopt a two-phase curriculum training strategy: In the first phase, we prioritize posterior convergence by optimizing only $\mathcal{L}_{Recon}$ and $\mathcal{L}_{KL}$, which ensures accurate motion intent capture and prevents posterior collapse caused by premature alignment to the prior. The second phase gradually introduces $\mathcal{L}_{PC}$, $\mathcal{L}_{VL}$ and $\mathcal{L}_{Aux}$ through weight warming to enable efficient knowledge distillation from posterior to prior. Furthermore, lightweight parameter adaptation with LoRA is implemented, which involves freezing the SigLIP encoder and training only low-rank increments. This design allows $z_{vl}$ to effectively fuse cross-modal semantics with motion trend logic, thereby enhancing semantic-motion intent adaptability. This design lays a robust foundation for intent-driven flow matching in the subsequent Basal Ganglionic Brain and motion tracking in the Cerebellar Brain.

\subsection{The design of the Basal Ganglionic Brain} 

The Basal Ganglionic Brain is designed to bridge low-frequency semantic-motion intent from the Neocortical Brain and high-frequency humanoid stable control from Cerebellar Brain. To solve low inference frequency and cumulative error of autoregressive VLA models, we adopt conditional flow matching for real-time continuous motion generation. It takes dual conditional inputs including \qty{10}{\Hz} semantic latent vector \(z_{vl} \in \mathbb{R}^{256}\) from the Neocortical Brain and 38-dimensional Unitree G1 robot state \(s_t \in \mathbb{R}^{38}\) including 3D root position, 6D orientation and 29 joint angles, outputting \qty{50}{\Hz} motion sequence chunks \(M_t = [m_{t,1}, \dots, m_{t,l}]\) with \(l=10\) and stride=2, which ensures temporal continuity and smoothness of motion.

A lightweight single-stream Gemma\cite{team2024gemma} decoder is adopted for flow matching. This unifies the modelling of the token sequence \([z_{\text{vl}}, s_t, m_t]\), while motion tokens are generated by projecting the noise-data inference \(x_t\) with time embeddings and the decoder exclusively outputs the vector field \(v_\theta\) for motion sequence tokens. During training, time steps \(t \sim \text{Beta}(\alpha=1.5, \beta=1)\) and Gaussian noise \(\varepsilon \sim \mathcal{N}(0,I)\) are sampled to construct the linear path \(x_t = t\cdot\varepsilon + (1-t)\cdot m_t\), with the objective to minimize the masked mean squared error \(\left\| v_\theta(x_t, t, z_{\text{vl}}, s_t) - (\varepsilon - m_t) \right\|^2\). For inference, the model operates at 10 Hz, generating action chunks of length 10 via reverse ODE integration (Euler/Heun solver) from \(t=1\) to \(t=0\). To reduce latency and improve smoothness, only the first five frames of each generated chunk are fed into the downstream Cerebellar Brain. By executing these 5 frames per 10 Hz inference cycle, the system achieves an effective high-frequency control rate of 50 Hz. Optional weight-based fusion can also be applied across these overlapping frames to further improve temporal consistency. 

\subsection{The design of the Cerebellar Brain}

The Cerebellar Brain acts as a robust motion tracking module, translating high-frequency motion sequence chunks into stable, physics-aware whole-body control. Inspired by the TWIST \cite{ze2025twist} architecture, we use a two-stage teacher-student RL framework to address the instability and conservative behavior common in single-stage policies. First, a privileged teacher policy learns smooth, coordinated whole-body movements via RL by utilizing future reference motions. Next, a deployable student policy learns from the teacher through a combination of RL and behavior cloning (BC), relying solely on real-time proprioceptive feedback and current reference frames.

After the initial pre-training on a physics-informed
motion dataset, we fine-tune the student policy on our own task-specific dataset. This step improves dynamic consistency and physical realism, helping to bridge the sim-to-real gap and ensuring the robot can reliably perform complex tasks.

\FloatBarrier 
\begin{figure*}[!t]
\centering
\vspace{-6pt}
\includegraphics[width=0.9\textwidth]{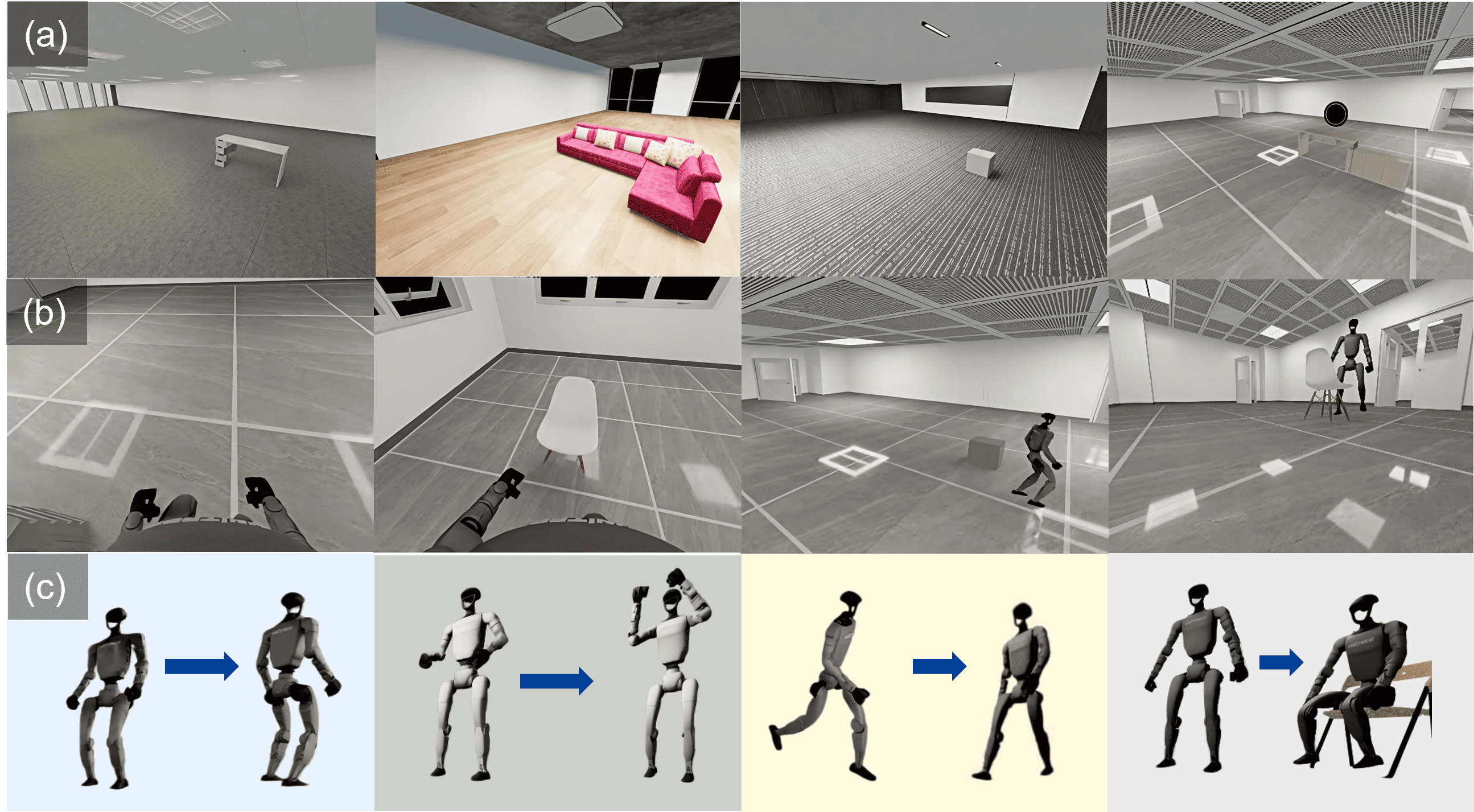}
\caption{\textbf{Visualization of the VLA dataset.} \textbf{(a)} Diverse visuals with various Scenes and Items. \textbf{(b)} Diverse camera angles with ego and exo views. \textbf{(c)} Diverse task from turning around to standing up}
\label{data_gen}
\vspace{-6pt}
\end{figure*}
\subsection{VLA Data Generation}  

In order to generate data for VLA humanoid whole-body control effectively, a series of practical methods were conducted in Isaac Lab \cite{Mittal_Isaac_Lab_-_2025}. Initially, we deployed Pico devices to remotely capture multiple motion sequences, including specific tasks such as sitting down on specified chairs, moving to a target location. Subsequently, the trained Cerebellar Brain's motion tracker is used to output physics-aware data that is dynamically consistent during actual execution. Furthermore, several scenarios were constructed in Isaac Lab, including laboratories and living rooms equipped with chairs, tables, and other relevant objects.

Within the Isaac Lab simulation environment, motion sequences captured from the experiment were replayed using Cerebellar Brain's motion tracker, while real-time images were captured from both first-person and third-person perspectives. In order to maximise data diversity, the objects and backgrounds in the simulation were automatically replaced during each replay of the motion sequences. The diversity of the VLA datasets can be seen in the Fig. \ref{data_gen}.
\newcommand{\ct}{\textbf{c}_t}
\newcommand{\bt}{\textbf{b}_t}
\newcommand{\vxcmd}{\textbf{v}_x^{\footnotesize{\textrm{cmd}}}}
\newcommand{\vycmd}{\textbf{v}_y^{\footnotesize{\textrm{cmd}}}}
\newcommand{\wzcmd}{\boldsymbol{\omega}_z^{\footnotesize{\textrm{cmd}}}}
\newcommand{\gaitparams}{\boldsymbol{\theta}^{\footnotesize{\textrm{cmd}}}}
\newcommand{\offsetcmd}{\boldsymbol{\theta}_1^{\footnotesize{\textrm{cmd}}}}
\newcommand{\phasecmd}{\boldsymbol{\theta}_2^{\footnotesize{\textrm{cmd}}}}
\newcommand{\boundcmd}{\boldsymbol{\theta}_3^{\footnotesize{\textrm{cmd}}}}
\newcommand{\freqcmd}{\boldsymbol{f}^{\footnotesize{\textrm{cmd}}}}
\newcommand{\durationcmd}{\boldsymbol{r}^{\footnotesize{\textrm{cmd}}}}
\newcommand{\heightcmd}{\boldsymbol{h}_z^{\footnotesize{\textrm{cmd}}}}
\newcommand{\pitchcmd}{\boldsymbol{\phi}^{\footnotesize{\textrm{cmd}}}}
\newcommand{\stancecmd}{\boldsymbol{s}_y^{\footnotesize{\textrm{cmd}}}}
\newcommand{\swingcmd}{\boldsymbol{h}_z^{f, \footnotesize{\textrm{cmd}}}}
\newcommand{\cfootcmd}{C^{\footnotesize{\textrm{cmd}}}_{\textrm{foot}}}
\newcommand{\ot}{\textbf{o}_{t}}
\newcommand{\oth}{\textbf{o}_{t-H...t}}
\newcommand{\cth}{\textbf{c}_{t-H...t}}
\newcommand{\bth}{\textbf{b}_{t-H...t}}
\newcommand{\ath}{\textbf{a}_{t-H-1...t-1}}
\newcommand{\timesth}{\textbf{t}_{t-H...t}}
\newcommand{\st}{\textbf{s}_{t}}
\newcommand{\timest}{\textbf{t}_{t}}
\newcommand{\at}{\textbf{a}_{t}}
\newcommand{\atp}{\textbf{a}_{t-1}}
\newcommand{\vt}{\textbf{v}^{\footnotesize{\textrm{cmd}}}_t}
\newcommand{\dt}{\textbf{d}_t}
\newcommand{\qt}{\textbf{q}_{t}}
\newcommand{\qdt}{\dot{\textbf{q}}_{t}}
\newcommand{\gt}{\textbf{g}_{t}}
\newcommand{\xt}{\textbf{x}_t}
\newcommand{\zt}{\textbf{z}_t}
\newcommand{\zht}{\hat{\textbf{z}}_t}
\newcommand{\Rnum}{\mathbb{R}}
\newcommand{\rwz}{r_{\omega^{\footnotesize{\textrm{cmd}}}_z}}
\newcommand{\rvx}{r_{v^{\footnotesize{\textrm{cmd}}}_{x,y}}}
\newcommand{\rcf}{r_{c^{\footnotesize{\textrm{cmd}}}_f}}
\newcommand{\rcv}{r_{c^{\footnotesize{\textrm{cmd}}}_v}}
\newcommand{\rbh}{r_{\heightcmd}}
\newcommand{\rp}{r_{\pitchcmd}}
\newcommand{\rst}{r_{\stancecmd}}
\newcommand{\rsw}{r_{\swingcmd}}

\section{EXPERIMENT}

\begin{table*}[htbp]
  \centering
    \renewcommand{\arraystretch}{1.2}
  \setlength{\tabcolsep}{6pt} 
  \small
    \caption{Comparison of core metrics across different ablation settings.Best results are viewed in \textbf{bold black}.}
  \begin{tabular}{@{}lccccccc@{}}
    \toprule
    Metric Name & Full & PC Off & Disc Off & VL Off & Lora Off & Textcond0 & Curr. Off \\ 
    \midrule
    Recon. Post & \textbf{0.017} & 0.020 & 0.019 & 0.020 & 0.020 & 0.041 & 0.081 \\ 
    Recon. Prior & \textbf{0.023} & 0.027 & 0.026 & 0.026 & 0.029 & 0.045 & 0.081 \\ 
    Future Shuffle Gap & \textbf{1.134} & 0.679 & 0.852 & 0.940 & 0.958 & 0.516 & 0.001 \\ 
    KL CVAE & \textbf{36.095} & 36.369 & 29.632 & 26.835 & 29.016 & 38.783 & 9.100 \\ 
    Retrieval (Cross Ep.) & \textbf{0.859} & 0.879 & 0.872 & 0.037 & 0.869 & 0.846 & 0.846 \\ 
    Retrieval (Same Task) & \textbf{0.357} & 0.364 & 0.365 & 0.086 & 0.362 & 0.349 & 0.352 \\ 
    Retrieval Top-1 & \textbf{0.357} & 0.363 & 0.361 & 0.016 & 0.361 & 0.349 & 0.350 \\
    Text-Use Gap (Zero) & \textbf{1.921} & 1.119 & 1.546 & 3.706 & 4.642 & 0.454 & 1.319 \\ 
    $z$-Use Gap & \textbf{0.921} & 1.030 & 0.885 & 0.723 & 0.763 & 0.784 & 3.397 \\ 
    \bottomrule
  \end{tabular}
  \label{tab:ablation_core_metrics}
\end{table*}

\subsection{The ablation of the Neocortical Brain}

A comprehensive ablation study was conducted to validate the contribution of each core component, with one component ablated at a time. All experimental settings including the training schedule, hyperparameters, data preprocessing and dataset splits were fixed to ensure a fair comparison. All models were evaluated with a batch size of 48 and 16 evaluation batches. Ablated models are evaluated across three dimensions:

\begin{itemize}[leftmargin=1.5em,
                  topsep=4pt,
                  itemsep=2pt,
                  parsep=0pt]
\item \textbf{Reconstruction fidelity}: Measured by Recon. Post and Recon. Prior, quantifying trajectory prediction accuracy under posterior- and prior-conditioned inference.
\item \textbf{Latent regularization \& information utilization}: Evaluated via z-Use Gap, Text-Use Gap (Zero), Future-Shuffle Gap, and KL-CVAE, measuring latent usage, conditioning effectiveness, posterior dependence on future information, and latent distribution alignment.
\item \textbf{Semantic alignment}: Assessed by Retrieval Top-1, Retrieval (Cross Ep.) and Retrieval (Same Task), quantifying latent-language matching in standard, cross-episode, and task-specific scenarios.
\end{itemize}

The specific ablation operations for each component are listed as follows:
\begin{itemize}[leftmargin=1.5em,
                  topsep=4pt,
                  itemsep=2pt,
                  parsep=0pt]
    \item \textbf{VL alignment (VL off):} Ablated by setting all VL and VL-cond weights to zero.
    \item \textbf{PC distillation (PC Off):} Ablated by clearing PC-related weights.
    \item \textbf{Disc regularization (Disc Off):} Ablated by turning off the discriminator.
    \item \textbf{LoRA adaptation (Lora Off):} Ablated by disabling LoRA for vision and text encoders.
    \item \textbf{Text conditioning (Textcond0):} Ablated by setting its scale to zero.
    \item \textbf{Two-phase curriculum strategy (Curr. Off):} Ablated by adopting a uniform loss weighting scheme during training.
\end{itemize}

As demonstrated in Table \ref{tab:ablation_core_metrics}, the ablation results reveal the contribution of each component. For VL Off, its ablation induces substantial retrieval collapse, evidenced by Retrieval Top-1 declining from 0.357 to 0.016 and Retrieval (Cross Ep.) from 0.859 to 0.037, which proves its essential role in semantic grounding. For Textcond0, ablating it significantly reduces the Text Use Gap (Zero) from 1.921 to 0.454 and degrades retrieval performance, verifying language’s function in shaping intent. PC Off causes minor retrieval changes but elevates reconstruction errors, drops Future Shuffle Gap and raises KL CVAE and z-Use Gap, degrading prior-posterior alignment. Disc Off barely affects metrics but reduces cross-modal performance by aligning vision-language and language-only latent spaces for scenarios without visual input. Lora Off leads to a moderate retrieval decline with consistent non-dominant gains in most metrics. Curr. Off drastically reduces Future Shuffle Gap to 0.001 and worsens reconstruction metrics, validating staged training. Overall, our full model balances semantic alignment, prior usability and latent stability, outperforming all ablated variants.

\subsection{The ablation of the Basal Ganglionic Brain}
\begin{figure}[!t]
    \centering
    \includegraphics[width=\columnwidth]{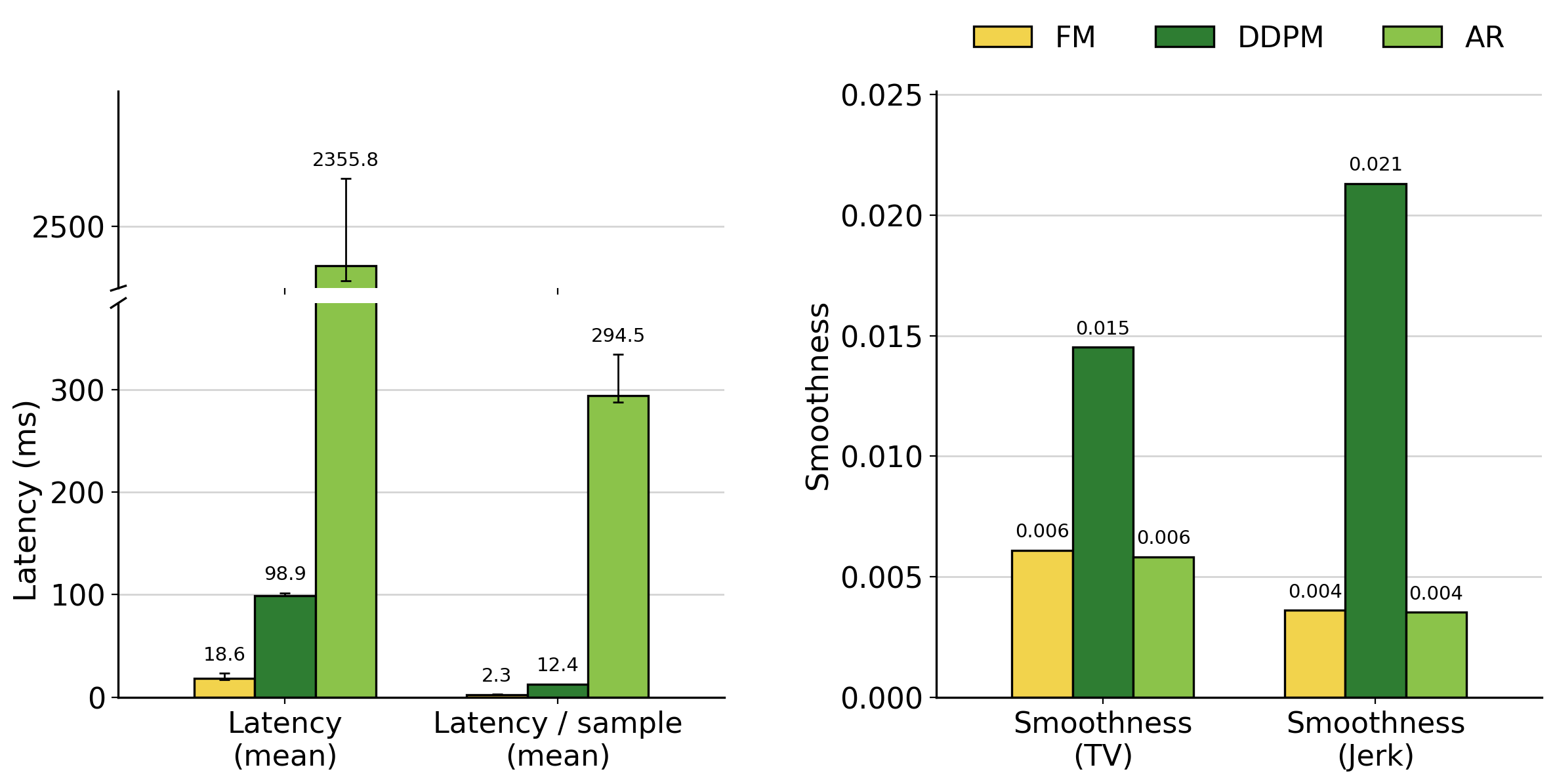}
    \caption{\textbf{Performance benchmarking of the Basal Ganglionic Brain.} The proposed flow-matching (FM) paradigm is evaluated against autoregressive (AR) and Denoising Diffusion Probabilistic Model (DDPM) baselines. }
    \label{fig:FM_benchmark}
\end{figure}

The inference efficiency and motion smoothness of the Basal Ganglionic Brain are evaluated against autoregressive (AR)\cite{intelligence2025pi_} and Denoising Diffusion Probabilistic Model (DDPM)\cite{ho2020denoising} baselines as follows:
\begin{itemize}
[leftmargin=1.5em,
                  topsep=4pt,
                  itemsep=2pt,
                  parsep=0pt]
\item \textbf{AR:} Generates motion sequences in a step-by-step autoregressive manner, with each frame dependent on the generation results of previous frames.
\item \textbf{DDPM:} Generates valid motion sequences by gradually denoising random noise signals through a multi-step reverse diffusion process.
\end{itemize}

As shown in Fig. \ref{fig:FM_benchmark}, a mean latency of 18.65 ms and per-sample latency of 2.33 ms are achieved by our flow-matching (FM) approach, which is 5.3 times faster than DDPM and 126 times faster than AR. In terms of motion smoothness, a total variation of 0.0061 and jerk of 0.0036 are yielded by FM, matching the high smoothness of AR while outperforming DDPM by a significant margin. These results demonstrate that the FM paradigm effectively resolves the trade-off between inference speed and motion quality, enabling real-time \qty{50}{\Hz} high-frequency motion generation for humanoid whole-body control.

\begin{figure*}[!t]
\centering
\vspace{-6pt}
\includegraphics[width=0.9\linewidth]{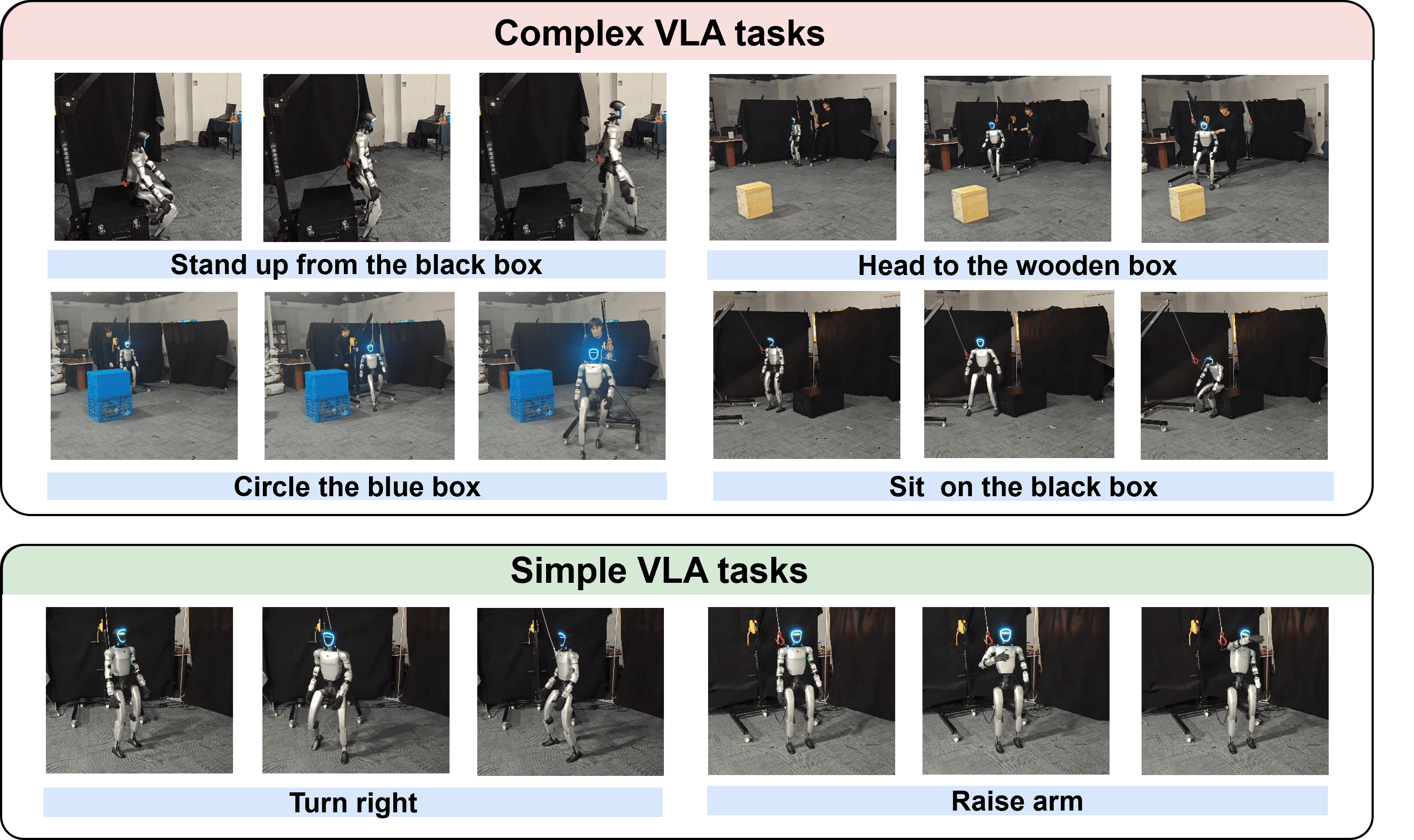}
\caption{\textbf{Real-world execution of semantically guided whole-body tasks by the Unitree G1 humanoid robot.} \textbf{Top:} Complex VLA maneuvers requiring continuous spatial navigation and dynamic multi-limb coordination. \textbf{Bottom:} Basic VLA tasks demonstrating responsive semantic execution and robust postural stability. These results validate the system's capacity to maintain physical compliance and dynamic consistency during unconstrained deployment.}
\label{actual_robot}
\vspace{-6pt}
\end{figure*}

\begin{table}[t] 
  \centering
  \vspace{-6pt}
  \setlength{\tabcolsep}{4pt}  
  \renewcommand{\arraystretch}{1.15}  
  
  \caption{Comparison of Success Rates (\%) between LeVERB and PhysiFlow. Best results are viewed in \textbf{bold black}.}
  \label{tab:main_results}
  
  \resizebox{\columnwidth}{!}{
  \begin{tabular}{@{}llcc@{}}  
    \toprule
    Task Type & Task Name & LeVERB~\cite{xue2025leverb} & PhysiFlow \\ 
    \midrule    
    \multirow{5}{*}{Complex} & Nav. (Long) & 31.2 & \textbf{63.6} \\
                             & Nav. (Short) & \textbf{78.5} & 71.7 \\
                             & Nav. \& Sit & 5.8  & \textbf{18.1} \\
                             & Nav. \& Circle & 54.5& \textbf{69.2} \\
    \midrule
    \multirow{4}{*}{Regular} & Stand up & 88.6 & \textbf{90.9} \\
                             & Locomotion & 97.2 & \textbf{100.0} \\
                             & turn around & 68.7 & \textbf{78.6} \\
                             & turn left & 81.3 & \textbf{81.8} \\
                             & raise arm & 79.1 & \textbf{100.0} \\
    \midrule
    
    Average & Overall & 65.0& \textbf{74.9} \\ 
    \bottomrule
  \end{tabular}
  }\label{simulation results}
\vspace{-10pt}
\end{table}

\subsection{Simulation Experiments}

To evaluate the proposed PhysiFlow framework, we conducted comprehensive simulation experiments using the Unitree G1 robot in Isaac Lab, with all inference executed on a single 48GB RTX 4090 GPU. We designed nine representative whole-body tasks, categorized into complex coordinated motions and regular basic tasks. LeVERB was selected as the primary baseline and locally reproduced within our environment to ensure a fair comparison. Task success rate serves as the core quantitative metric.

Table \ref{tab:main_results} presents the experimental results. While the reproduced baseline performs well on basic motions and short-distance navigation, PhysiFlow outperforms it in most categories, especially in tasks requiring continuous coordination. Overall, PhysiFlow achieves an average success rate of 74.9\%, exceeding LeVERB's 65.0\% by 9.9 percentage points. The proposed framework delivers major improvements in complex tasks. For example, it increases the success rate of the ``Navigation \& Circle'' task from 54.5\% to 69.2\%, and more than doubles it for ``Nav. (Long)''  from 31.2\% to 63.6\%. These results clearly demonstrate that our multi-brain VLA architecture successfully combines semantic guidance with fast inference and stable motion.

\subsection{Real-World Experiments}

We deploy PhysiFlow on the physical Unitree G1 humanoid robot to validate real-world feasibility, motion stability, and semantic consistency. In this paper, we evaluate the hardware performance via open-loop execution of generated motion sequences on the robot, which provides a controlled and reproducible setting for assessing sim-to-real transfer of whole-body behaviors.

As shown in Fig. \ref{actual_robot}, the Unitree G1 robot completed various semantically guided tasks via the PhysiFlow architecture, proving its reliable sim-to-real transfer ability. The system achieved smooth movements, strong limb coordination in complex VLA tasks and excellent dynamic stability for basic commands, which confirms our flow-matching paradigm generates physically consistent motion sequence chunks. More importantly, PhysiFlow effectively bridges the divide between high-level multimodal semantic reasoning and low-level physical execution.

\section{CONCLUSIONS}

The paper introduces PhysiFlow, a physics-aware policy for humanoid whole-body VLA control, with a bio-inspired multi-brain architecture. This architecture comprises the Neocortical Brain, which fuses semantic-motion intent via curriculum-based CVAE, the Basal Ganglionic Brain, which generates high-frequency motion sequences through latent vector-driven flow matching, and the Cerebellar Brain, which enforces physics-aware constraints via joint fine-tuning. By decoupling high-level semantic inference from low-level motion generation and stable tracking, the proposed framework resolves the trade-off between inference efficiency, semantic generalization, and dynamic balance in humanoid VLA systems. Experiments on Unitree G1 robots demonstrate significant improvements in task success rates and motion stability, and ablation studies validate the contribution of each core component. Future work may involve integrating world models and VLA into full-body control for humanoid robots, thereby addressing current challenges such as limited datasets, jitter and low success rates.

\section*{APPENDIX}

\subsection{Sensitivity Analysis of Latent Dimension}
\begin{figure}[!t]
    \centering
    \includegraphics[width=\columnwidth]{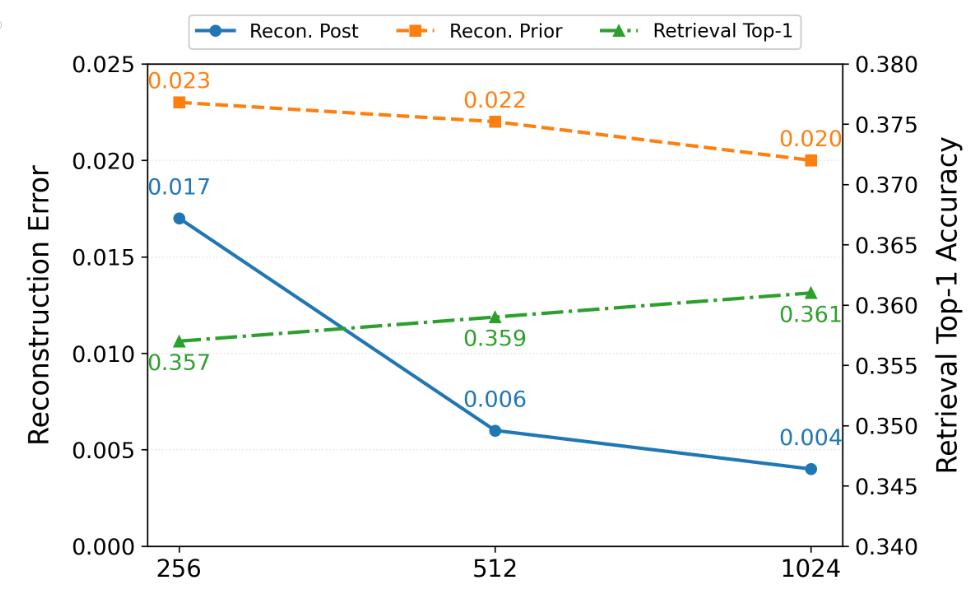}
    \caption{\textbf{Performance metrics across different latent dimensions including 256, 512 and 1024.} The left axis tracks the posterior and prior reconstruction errors, while the right axis displays the Top-1 retrieval accuracy.} \label{fig:latent}
\end{figure}

The dimensionality of the semantic latent vector $z_{vl}$ directly affects the representation capability and inference efficiency of our multi-brain framework. To determine the optimal configuration, we conducted a sensitivity analysis comparing three candidate dimensions.

As shown in Fig. \ref{fig:latent}, increasing the latent dimension yields no significant improvement in core metrics including reconstruction error and retrieval accuracy. Meanwhile, higher dimensions introduce extra parameter overhead, increase memory consumption, and reduce inference speed. Importantly, restricting the dimension is not a compromise for low-cost computational solutions, but a strict necessity to meet the high-frequency inference requirements of humanoid robots. We thus select the 256-dim configuration for the final design, achieving the optimal balance between performance and real-time execution.

\bibliographystyle{ieeetr}

% \footnotesize{
% \bibliography{arxiv}}

\end{document}